\documentclass[runningheads]{llncs}
\usepackage[T1]{fontenc}
\usepackage{graphicx}
\usepackage{color}
\usepackage{hyperref}

\usepackage{mathrsfs, mathtools, amssymb}
\usepackage{comment}
\usepackage{color}
\usepackage{tikz}
\usepackage{booktabs}
\usepackage{nicematrix}
\usepackage{multirow}
\usepackage{enumitem}
\usepackage{caption}
\usepackage{wrapfig}

\usepackage{biblatex} 
\begin{document}
\title{Discrete World Models via Regularization}
\titlerunning{Discrete World Models via Regularization}
% If the paper title is too long for the running head, you can set an abbreviated paper title here
%

\author{Davide Bizzaro\inst{1,2}\orcidID{0009-0004-0420-0453} \and Luciano Serafini\inst{2}\orcidID{0000-0003-4812-1031}}
\authorrunning{D. Bizzaro and L. Serafini}
% First names are abbreviated in the running head.
% If there are more than two authors, 'et al.' is used.
%
\institute{University of Padua, Padua, Italy \\
\email{davide.bizzaro@phd.unipd.it}\\
\and
Fonadazione Bruno Kessler, Trento, Italy \\
\email{serafini@fbk.eu}}

\maketitle              % typeset the header of the contribution

\begin{abstract}
%Discrete world models offer powerful abstractions for planning in combinatorial domains. 
%Prior work typically uses decoder-based reconstruction, or else contrastive or supervised signals.
World models aim to capture the states and dynamics of an environment in a compact latent space. %to enable prediction and planning.
Moreover, using Boolean state representations is particularly useful for search heuristics and symbolic reasoning and planning.
Existing approaches keep latents informative via decoder-based reconstruction, or instead via contrastive or reward signals.
In this work, we introduce Discrete World Models via Regularization (DWMR): a reconstruction-free and contrastive-free method for unsupervised Boolean world-model learning. 
In particular, we introduce a novel world-modeling loss that couples latent prediction with specialized regularizers. Such regularizers maximize the entropy and independence of the representation bits through variance, correlation, and coskewness penalties, while simultaneously enforcing a locality prior for sparse action changes. To enable effective optimization, we also introduce a novel training scheme improving robustness to discrete roll-outs.
Experiments on two benchmarks with underlying combinatorial structure show that DWMR learns more accurate representations and transitions than reconstruction-based alternatives.
Finally, DWMR can also be paired with an auxiliary reconstruction decoder, and this combination yields additional gains.

  %\and Neurosymbolic planning \and Boolean latent states}
\end{abstract}

\section{Introduction}
%We want to develop a neuro-symbolic world model that can be trained end-to-end from unsupervised traces, to produce propositional representations of the states of an environment, where actions are taken. We suppose the environment to be fully-observable and deterministic, apart for noise in the perceptions; and we suppose the set of actions to be discrete and known. The focus is on building a good transition model, and no reward function is provided.
World models %and State Representation Learning 
aim to capture the states and dynamics of an environment in a compact latent space, enabling prediction, planning and downstream control from high-dimensional observations \cite{lesort2018state, ha2018world}. %In neurosymbolic settings, a key goal is to obtain propositional representations that expose a discrete structure while remaining trainable end-to-end from raw experience. 
Within this setting, we specifically target \emph{discrete} latent states made of bits that collectively form a propositional description of the environment state. Such discrete encodings are particularly appealing for planning and combinatorial domains, since they can mirror the underlying symbolic state space (e.g., board configurations in puzzles), allow exact state re-identification via bitwise equality, reduce model drift, and enable strong search heuristics and faster policy adaptation \cite{agostinelli2024deepcubeai,meyer2023harnessing}.
%and naturally support metrics such as Hamming distance for search and heuristics design. In fact, recent work on discrete world models for heuristic search has shown that discrete latents can improve long-horizon planning by reducing model drift and enabling stronger heuristics \cite{agostinelli2024deepcubeai,meyer2023harnessing}. %, reinforcing the view that symbolic or propositional structure is a natural interface between learned models and classical search.

%Typically, models addressing similar objectives include a generative component, allowing the reconstruction of the perception from the latent representation. This reconstruction/autoencoding objective is there to prevent latent collapse — a situation where the encoder produces trivial representations that are highly predictable — by enforcing that the latent variables retain sufficient information about the input. However, in our approach, we deliberately avoid including such a generative part, aiming instead to rely purely on the predictive and regularization losses to structure the propositional latent space.
Most existing approaches with similar objectives include a generative decoder that reconstructs observations from latent states \cite{ha2018world,asai2022classical,agostinelli2024deepcubeai,hafner2025dreamerv3}. This is used to provide a reconstruction feedback nudging the latent representation to retain information about the input, and thus prevent \emph{latent collapse}: the situation where the learned representations are predictable from the previous, yet trivial and non-informative. However, pixel-level reconstruction may focus more on noisy perceptual details than on what is relevant for predicting actions' effects \cite{lesort2018state,deng2022dreamerpro,burchi2024mudreamer}. 
%A complementary line of work relies on contrastive objectives, which bring ``positive'' pairs (e.g., augmented views) close and push ``negative'' pairs apart \cite{oord2018representation,okada2020dreaming}. 
%More recently, non-contrastive joint-embedding methods have shown that collapse can also be avoided without explicit negatives, by relying on architectural asymmetry and explicit regularizers \cite{grill2020bootstrap,zbontar2021barlow,bardes2022vicreg,bardes2024vjepa}. However, these methods still build on augmented views of the same underlying instance, instead of action-conditioned world modeling.
Another line of work, particularly within the context of unsupervised learning with augmented views and continuous representations rather than discrete world models, employs alternative strategies to prevent latent collapse \cite{lecun2022path}: contrastive learning \cite{oord2018representation,okada2020dreaming}, architectural asymmetry \cite{grill2020bootstrap,bardes2024vjepa} and regularization \cite{zbontar2021barlow,bardes2022vicreg}.

In this work, we propose Discrete World Models via Regularization (DWMR), employing regularizations specifically tailored to world modeling and to Boolean latent representations. In particular, committing to Boolean states offers a most-informative distribution ---independent $Bernoulli(0.5)$ bits--- that provides a principled reference for shaping the latent space. Furthermore, this discreteness enables a natural structural prior, where actions are expected to flip only a sparse subset of bits. This motivates our two guiding research questions: \emph{(i) leveraging this natural ideal of factorized bits, can we design a regularization scheme that yields informative, non-collapsed Boolean representations in a world model?} and \emph{(ii) is such a tailored regularization strong enough to dispense with pixel-level decoders or contrastive machinery altogether?}

We consider this problem in fully observable, deterministic environments (up to perceptual noise), with a known discrete action set and no reward function. Our focus is exclusively on learning a good representation and a good transition model in latent space, decoupled from policy learning. 
%To investigate this, we consider a model consisting of an encoder that maps each image to a Boolean latent vector, and an action-conditioned predictor that takes the current latent state and the executed action as input to predict the next latent state. Training is driven by a loss that seeks to:
%This leads to our central research questions: %\emph{can informative neurosymbolic structure emerge purely from reward-free predictive constraints in latent space, without  pixel-level reconstruction, contrastive supervision and augmented-views?}
%\emph{(i) can informative discrete structure emerge purely from reward-free forward dynamics in latent space, without pixel-level reconstruction, contrastive supervision, or augmented views?} and \emph{(ii) does committing to Boolean latent states provide more principled regularization targets than in the continuous case, by leveraging the natural ideal of factorized bits?}
%The model we propose is made of an \emph{encoder} that produces a Boolean vector from an image, and of a \emph{predictor} that takes as input the current latent state (outputted by the encoder) plus the action taken, and tries to produce the next latent state. 
%The training is done with loss functions that seeks to:
To investigate this, we consider a model consisting of an \emph{encoder} that maps each image to a Boolean latent vector, and a \emph{predictor} that takes the current latent state and the executed action as input and predicts the next latent state.

Concretely, we introduce a Boolean world-modeling loss that couples action-conditioned prediction with regularizers enforcing informative, factorized, and sparsely changing bit codes. 
Such loss function is designed to simultaneously:
%Training is driven by a loss that seeks to:
\begin{itemize}%[label = \arabic*)]
    \item minimize the prediction error of the next latent state,
    \item maximize (a proxy for) the entropy of each representation bit, \label{it:var}
    \item maximize (a proxy for) the independence of the representation bits,
    \item impose a locality prior on actions' effects, favoring transitions that flip only a small number of bits. %minimize a regularization term favoring the locality of actions (meaning that each step should change only few representation bits).
\end{itemize}
Together, these components yield a collapse-resistant training objective that is explicitly aligned with the structure and goals of Boolean world modeling, enabling a compact, meaningful latent space without relying on pixel reconstruction, contrastive pairs, or augmented views.

\section{Related Works}
Our work lies in the context of world models and state representation learning \cite{lesort2018state}, with a focus on discrete latent spaces. It is related to model-based reinforcement learning, neurosymbolic planning, and joint-embedding methods.

\subsection{Discrete World Models}
Model Based Reinforcement Learning (MBRL) seeks to learn a world model to roll-out imagined future states for decision making. Among the most relevant approaches in this area is the \textit{Dreamer} series \cite{hafner2019dreamer, hafner2020dreamerv2, hafner2025dreamerv3}. While \textit{DreamerV2} \cite{hafner2020dreamerv2} introduced categorical latents akin to our discrete focus, these models rely on generative decoders to reconstruct pixels, which can prioritize visual fidelity over control-relevant dynamics. To address this, \textit{reconstruction-free} approaches have emerged. Methods like \textit{Dreaming} \cite{okada2020dreaming} and \textit{DreamerPro} \cite{deng2022dreamerpro} replace pixel loss with contrastive objectives or prototype learning, respectively. Others, such as \textit{MuDreamer} \cite{burchi2024mudreamer} and \textit{TD-MPC2} \cite{hansen2024tdmpc2}, rely heavily on predicting values or rewards to shape the latent space. Unlike these approaches, we avoid the reward-and-policy loop, focusing on unsupervised, goal-independent learning.

A related line of work aims at bridging subsymbolic perception with symbolic planning/search in  discrete latent spaces. %\textit{DeepCubeA} \cite{agostinelli2019deepcubea} and the subsequent 
\textit{DeepCubeAI} \cite{agostinelli2024deepcubeai} combines MBRL with heuristic search on the learned discrete latents; \textit{LatPlan}~\cite{asai2022classical} induces a propositional action model from unlabeled transitions; \textit{VAE-IW} \cite{dittadi2021atari} uses width-based search on learned discrete representations, but the roll-outs are given by an external simulator, not predicted in imagination.
We too represent latent states as explicit bit vectors amenable to planning and search. However, our transitions are directly executable in imagination, unlike \textit{Latplan} and \textit{VAE-IW}, plus we avoid the generative component and use regularization instead. %Moreover, we ensure that the state representation emerges directly from the forward dynamics in the latent space, rather than being driven by a generative bottleneck.

%In combinatorial domains, DeepCubeA~\cite{agostinelli2019deepcubea} learns a heuristic from high-dimensional inputs and solves Rubik's Cube with A* search, while DeepCubeAI~\cite{agostinelli2024deepcubeai} extends this to a discrete world model with a goal-conditioned heuristic that plans directly in latent bit space. 
%In continuous control environments, \cite{umili2021symbolic} jointly learns symbol grounding, a symbolic transition model, and a value function from interaction, and performs online symbolic planning for action selection.

\subsection{Joint-Embedding Predictive Architectures (JEPA)}
In contrast to the previous literature, Joint-Embedding Predictive Architectures (JEPA) \cite{lecun2022path} learn representations by predicting the embeddings of future or masked views without rewards and pixel-level decoding. While contrastive methods (e.g., \textit{CPC} \cite{oord2018representation}) prevent collapse via negative pairs, non-contrastive approaches like \textit{BYOL} \cite{grill2020bootstrap}, \textit{Barlow Twins} \cite{zbontar2021barlow} and 
\textit{VICReg} utilize architectural asymmetry or variance-covariance regularization to maintain information content. 
Recent works like \textit{I-JEPA} \cite{assran2023ijepa} and \textit{V-JEPA} \cite{bardes2024vjepa} extend this to image and video masking; however all these works rely on augmented views and and not on action-condition dynamics. A work focusing on such dynamics is \textit{DINO-WM} \cite{zhou2024dinowm}, which however uses a frozen pre-trained encoder, and does not learn the state representation.   

%Our work adapts JEPA principles to an action-conditioned setting, %without data-augmentation, %and uses regularization terms specifically designed for boolean codes. Unlike \textit{DINO-WM}, and, unlike \textit{DINO-WM}, the encoder is trained end-to-end so that the bit semantics co-adapt with the transition predictor. 
Our work adapts JEPA principles to action-conditioned world modeling while learning the encoder end-to-end.  
Furthermore, instead of contrastive losses and masking data augmentation, we regularize directly for informative, factorized binary codes, with a locality bias on actions effects. This is analogous to the variance and covariance terms in \textit{VICReg} \cite{bardes2022vicreg}, but extended and explicitly formulated for Boolean instead of continuous encodings.
\section{Architecture and Training}

\begin{figure}[ht]
  \centering
  \includegraphics[width=\linewidth]{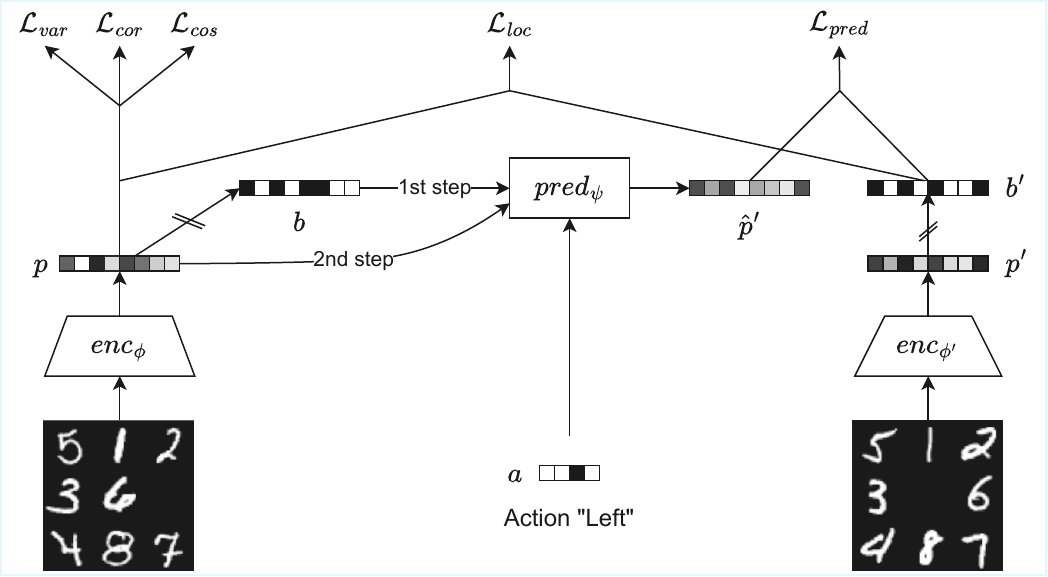}
    \caption{Overview of the model architecture and of the loss function. Encoders map successive observations into a shared Boolean latent space, and a predictor transforms the current latent state into the next, given the action. We illustrate and evaluate this setup on an 8-puzzle benchmark with MNIST digits, where actions move the blank tile and induce local state changes and new renderings of the digits. The crossed lines denote operations that stop the gradient, and we introduce a training scheme with two-steps updates: initially, the predictor is updated solely based on hard bits $b$, followed by a joint update of the encoder and predictor using probabilities $p$. Test-time inference relies only on the pathway involving $b$. The parameters $\phi'$ are an EMA copy of the parameters $\phi$.
}
  \label{fig:model}
\end{figure}

%Our model maps two successive views through separate encoders into the same discrete latent space, trains a predictor to transform the first into the second (given also the action taken), and optimizes so predicted and true target representations are close. % learning abstract predictive features without reconstructing pixels.
DWMR implements an action-conditioned joint-embedding predictive architecture with discrete latent space (see Fig.~\ref{fig:model}). Given two successive observations, an encoder $enc_\phi$ maps each observation to a Boolean latent vector, producing a current and a target latent state. A predictor $pred_\psi$ then takes the current latent state together with the executed action and outputs a prediction for the next latent state. 
Training jointly optimizes the parameters $\phi$ and $\psi$ so that the predicted representation closely matches the target one, while additional regularization encourages the learned Boolean codes to be informative, and to have sparse changes under each action. Indeed, what distinguishes DWMR is the design of a loss function that reliably induces informative discrete latent representations and dynamics, even without the reconstruction feedback from a generative component. 

\begin{comment}
Our model implements an action-conditioned joint-embedding predictive architecture with a discrete latent space. Given two successive observations $x, x'$ and the action $a$ taken between them, an encoder
$enc_\phi : \mathcal{X} \rightarrow \{0,1\}^K$ maps the observation to Boolean latent vectors
$b = enc_\phi(x)$ and $b' = enc_\phi(x')$.
A predictor $pred_\psi : \{0,1\}^K \times \mathcal{A} \rightarrow [0,1]^K$
then takes the current latent state and the executed action, and outputs a prediction $\hat{b}' = pred_\psi(b, a)$ for the next latent state.
Training jointly optimizes $\phi$ and $\psi$ so that the predicted representation $\hat{b}'$ matches the target representation $b'$, while additional regularization encourages the learned Boolean codes to be informative, and to vary only locally under each action.
\end{comment}

\subsection{Architecture}
The encoder $enc_\phi$ is a neural network with sigmoid outputs,  producing vectors of Bernoulli
bit-probabilities $p \in [0,1]^K$ and corresponding hard bits $b \in \{0,1\}^K$ obtained with a $0.5$ threshold.
The predictor $pred_\psi$ receives the current probabilities $p$ concatenated with a one-hot action $a$ and is implemented as another neural network yielding predicted soft values $\hat{p}' \in [0,1]^K$ for the next step. The corresponding predicted bits $\hat{b}' \in \{0,1\}^K$ are obtained by thresholding $\hat{p}'$ at $0.5$, and are used whenever a discrete representation is required.  
%The predictor $pred_\psi$ receives the current probabilities concatenated with a one-hot action $a$ to produce the predictions $\hat{b'}$ for the next step.

The target bits $b'$ used during training are obtained from a \emph{detached exponential moving average copy} $enc_{\phi'}$ of the encoder applied to the next image $x'$. Concretely, the parameters $\phi'$ of the target encoder are updated after each optimizer step as an exponential moving average (EMA) of the online encoder parameters $\phi$:
\[
\phi' \leftarrow \tau \, \phi' + (1 - \tau) \, \phi ,
\]
where $\tau \in (0,1)$ is the EMA coefficient. 
This mechanism is analogous to the method used in \emph{BYOL}~\cite{grill2020bootstrap}, where the slowly updated target network serves to avoid representation collapse. In our case, collapse is prevented instead by the loss function, and the EMA target encoder is used primarily to stabilize training. While \emph{BYOL} typically employs very slow EMA updates with
$\tau > 0.996$, we use a substantially smaller value (around
$\tau \approx 0.9$), which we found empirically to yield better predictive performance in our world-model setting.

\subsection{Loss Function}
%While a prediction loss ensures that the predictor outputs a next latent state consistent with the encoder’s representation of the subsequent observation, this objective alone is insufficient for learning a useful world model. Without additional constraints, the encoder may collapse to trivial or arbitrary codes, and the predictor may rely on degenerate shortcuts that fail to capture the underlying dynamics of the environment. We therefore introduce supplementary loss terms that shape the Boolean latent space itself: encouraging codes to be informative rather than collapsed, promoting decorrelation across latent dimensions, and enforcing locality so that each action produces only sparse, meaningful changes in the latent representation. These regularizers guide the model toward discovering a stable latent dynamics, more consistent with the true environment transitions.

A prediction loss enforcing consistency between predicted and encoded next states is not sufficient by itself to yield a useful world model: the predictor may exploit degenerate solutions that ignore the true dynamics of the environment, and the encoder can learn to generate non-informative codes. 
To avoid this, we add loss terms that explicitly shape the Boolean latent space: they encourage informative, non-collapsed codes, reduce redundancy across dimensions, and enforce that each action induces only sparse changes in the representation. %Together, these regularizers guide the model toward a informative and consistent latent dynamics.

Concretely, we propose a loss function which is a weighted sum of (i) a \emph{prediction} loss $\mathcal{L}_{\text{pred}}$ between predictor outputs and target bits,
(ii) \emph{variance}, \emph{correlation} and \emph{coskewness} regularizers $\mathcal{L}_{\text{var}}$, $\mathcal{L}_{\text{cor}}$ and $\mathcal{L}_{\text{cos}}$ to avoid representation collapse and maximize information content, and
(iii) a \emph{locality} regularizer $\mathcal{L}_{\text{loc}}$ to bias actions to flip only a few bits:
\begin{equation}\label{eq:loss}
   \mathcal{L}_{\text{DWMR}} \, \coloneqq \, \mathcal{L}_{\text{pred}}+\;\lambda_{\text{var}}\mathcal{L}_{\text{var}}+\lambda_{\text{cor}}\mathcal{L}_{\text{cor}}+\lambda_{\text{cos}}\mathcal{L}_{\text{cos}}+
   \lambda_{\text{loc}}\mathcal{L}_{\text{loc}}.
\end{equation}

%Given a minibatch of size $N$, we denote by $p \in [0,1]^{N \times K}$ the matrix of encoder Bernoulli probabilities over the first images in the batch. We also define the batch-standardized version
In what follows we describe each term that composes this loss function. To this end, we first fix some notation. Let $N$ denote the number of samples in a minibatch and $K$ the number of bits in the latent representation. We denote by $p \in [0,1]^{N \times K}$ the matrix of encoder Bernoulli probabilities over the first images in the batch. We then define the batch-normalized probabilities
\begin{equation}
    \tilde{p}_{mk} \coloneqq \frac{p_{mk} - \mathbb{E}_n[p_{nk}]}{\sigma_n[p_{nk}] + \epsilon},
\end{equation}
where %expectations and standard deviations are computed over the batch index $n$, and 
$\epsilon = 10^{-6}$ is a constant added for numerical stability. %All regularization terms introduced below are computed from $p$ and its standardized counterpart $\tilde{p}$.

\subsubsection{Prediction Loss $\mathcal{L}_{\text{pred}}$.} As prediction loss $\mathcal{L}_{\text{pred}}$, we use the binary cross-entropy (BCE) between the predicted probabilities $\hat{p}'$ outputted by $pred_\psi$ and the (detached) target bits $b'$ produced by $enc_{\phi'}$ over the next image:
\begin{equation}
    \mathcal{L}_{\text{pred}} \coloneqq \text{BCE}(\hat{p}', b') %= BCE(pred_\psi(enc_{\phi}(x), a), round(enc_{\phi'}(x')))
\end{equation}

\subsubsection{Variance Regularizer $\mathcal{L}_{\text{var}}$.}
To prevent degenerate codes where each bit is constant, we want to encourage \emph{per-bit variability}. %Given encoder probabilities $p \in [0,1]^{B\times K}$ over the first images in a batch of data, we compute the batch variance 
Let $v_k \coloneqq \mathrm{Var}_n[p_{nk}]$ be the batch variance for each bit $k$. The $\mathcal{L}_{\text{var}}$ penalty softly enforces a standard deviation per bit of at least $\gamma$:  
\begin{equation}
\mathcal{L}_{\text{var}}
\coloneqq \frac{1}{K}\sum_{k=1}^K
\max\!\Big(0,\; \gamma - \sqrt{v_k + \epsilon}\Big).
\end{equation}
Here $\gamma$ is a hyperparameter close to but lower than the $Bernoulli(0.5)$'s standard deviation of $0.5$, and $\epsilon = 10^{-6}$ is added for numerical stability. This hinge form of the loss yields zero cost when the standard deviation is high enough ($\sqrt{v_k + \epsilon}>\gamma$) and penalizes lower variability otherwise.

\subsubsection{Correlation Regularizer $\mathcal{L}_{\text{cor}}$.}
To promote independence across bits, we penalize off-diagonal absolute correlation among the encoder outputs.
%Let $\tilde{p} \coloneqq (p - \mathbb{E}_b[p])/ \sigma_b[p]$ be the standardized  and
So if $C \coloneqq \frac{1}{B-1}\tilde{p}^\top\tilde{p}$ is the correlation matrix, we minimize its mean absolute off-diagonal value:
\begin{equation}
\mathcal{L}_{\text{cor}}
\coloneqq \frac{1}{K(K-1)} \sum_{i \ne j} |C_{ij}|.
\end{equation}
%This drives redundant co-activations down while leaving the diagonal (per-bit variance) to be handled by $\mathcal{L}_{\text{var}}$. 
Differently from \textit{VICReg} \cite{bardes2022vicreg}, we found better results using the \emph{correlation} matrix instead of the \emph{covariance} matrix, i.e., dividing each dimension $k$ by its standard deviation along the batch $\sigma_n[p_{nk}] + \epsilon$.

\subsubsection{Coskewness Regularizer $\mathcal{L}_{\text{cos}}$.}
While $\mathcal{L}_{\text{cor}}$ suppresses \emph{pairwise} dependencies, higher-order co-activations can persist.
To discourage such occurrencies, we penalize standardized \emph{third-order} cross-moments (a.k.a.\ coskewness) of the bit-probabilities. 
%Let $p \in [0,1]^{B \times K}$ be the encoder probabilities in a batch, and $\tilde{p} = (p - \mathbb{E}_b[p])/ \sigma_b[p]$ their batch-normalized version.
Thus, we form the third-order moment tensor $
M_{ijk} = \mathbb{E}_n\!\left[\tilde p_{ni}\,\tilde p_{nj}\,\tilde p_{nk}\right]$ (using Einstein notation), 
and restrict to triplets with all indices distinct. %$\mathcal{S}=\{(i,j,k)\mid i\neq j,\; j\neq k,\; i\neq k\}$.
The loss is the mean absolute value of these cross-moments:
\begin{equation}
\mathcal{L}_{\text{cos}}
\coloneqq
\frac{1}{K(K-1)(K-2)}\, \sum_{i\neq j,\; j\neq k,\; i\neq k} \left| M_{ijk} \right|.
\end{equation}
$\mathcal{L}_{\text{cos}}$ complements $\mathcal{L}_{\text{cor}}$ by suppressing structured triplet interactions that would otherwise survive pairwise decorrelation. The cubic complexity was perfectly tractable in our setting, but in case one may reduce it by computing the loss only on some randomly sampled triplets $(i,j,k)$. 

\begin{comment}
\subsubsection{Locality Regularizer $\mathcal{L}_{\text{loc}}$.}
We want each action to change only a few bits---enough to capture a local, sparse transition
in the propositional state.
We therefore compute a soft Hamming distance between the encoder probabilities at successive timesteps:
\[
d_b = \frac{1}{K}\sum_{k=1}^K |p_{bk} - b'_{bk}| \ \mathbb{I}_{|p_{bk} - b'_{bk}|>0.5}.
\]
We then penalize deviations from a target window $[L,U]$:
\begin{equation}
\mathcal{L}_{\text{loc}}
= \mathbb{E}_b\Big[\max\!\big(0,\; |d_b - m| - w\big)^2\Big],
\quad
m = \frac{L+U}{2K},\;
w = \frac{U-L}{2K}.
\end{equation}
For example, if we use $L=1$ and $U=6$, the model is nudged to flip
at least one and at most six bits per move.
\end{comment}

\subsubsection{Locality Regularizer $\mathcal{L}_{\text{loc}}$.}
In most domains, the underlying dynamics are \emph{local} in a propositional state space: a single action typically modifies only a small subset of the state variables (e.g., the position of one tile in a puzzle or the angle of one joint in a robot) while leaving the remainder of the environment state invariant.
If the Boolean latent vector is to play the role of such a propositional state, it is natural to bias the model toward transitions that flip only few bits at each time step. This locality bias encourages the model to disentangle independent factors of variation and discover a representation where actions have local and possibly more interpretable effects.

Concretely, let $p_{nk} \in [0,1]$ be the probability of bit $k$ for sample $n$ produced by the encoder $\mathrm{enc}_{\phi}(x)$ at the current image $x$, and let $b'_{nk} \in \{0,1\}$ be the corresponding hard bit obtained from the target encoder $\mathrm{enc}_{\phi'}(x')$ applied to the next image $x'$. To quantify how many bits are effectively flipped by one transition, we define a soft Hamming distance between the current probabilities and the next-step
target bits for each sample $n$:
\begin{equation}
d_n \coloneqq \frac{1}{0.75 K} \sum_{k=1}^K 
\bigl| p_{nk} - b'_{nk} \bigr|\,
\mathbb{I}\!\left( \bigl| p_{nk} - b'_{nk} \bigr| > 0.5 \right),
\end{equation}
where $\mathbb{I}(\cdot)$ is the indicator function. 
A dimension contributes only when the difference crosses the $0.5$ threshold, meaning the non-zero terms in the sum are strictly within the range $(0.5, 1]$. We therefore normalize by $0.75K$ (instead of $K$) to account for the average magnitude of these active terms, obtaining a more accurate estimate of the fraction of flipped bits.

We then penalize deviations of $d_n$ from a desired window $[L, U]$ of flipped bits per action:
\begin{equation}
\mathcal{L}_{\text{loc}}
\coloneqq \mathbb{E}_n\Big[\max\!\big(0,\; |d_n - m| - w\big)^2\Big],
\quad
m = \frac{L+U}{2K},\;
w = \frac{U-L}{2K}.
\end{equation}
For example, if we use $L=1$ and $U=6$, the model is nudged to flip (approximately) at least one and at most six bits per move.

\subsection{Training Procedure}\label{sub:training}
\label{sec:training}

%We train the model under the usual constraint that backpropagation requires continuous values, while at test time we care about the \emph{discrete} latent transition function operating on binarized codes. If the predictor were trained only on probabilities, it could exploit information that disappears after thresholding; if it were trained only on hard bits, gradients could not flow reliably through the encoder.
A challenge in learning discrete latent dynamics is the tension between optimization and representation. On the one hand, we rely on backpropagation, which requires differentiable quantities. On the other hand, our transition model will ultimately operate on discrete Boolean states, and its performance at test time depends on its ability to consume and predict binarized latent codes rather than real-valued continuous vectors. 

To reconcile these requirements, we use a simple two-step training scheme on each minibatch. 
Let $p$ be the encoder probabilities, $b  \coloneqq \mathbb{I}({p\geq 0.5})$ the corresponding hard bits, and $\hat{p}'$ the predictor output targeting $b'$. The two steps are the following:

\begin{enumerate}
    \item \textbf{Predictor update on discrete inputs.} We first detach the encoder and update only the predictor parameters $\psi$ using the prediction loss $\mathcal{L}_{\text{pred}}$. In this step, the predictor is trained with inputs $b$ and targets $b'$. Thus, the predictor is taking discrete codes as input, which is exactly the regime it will face at test time.
    \item \textbf{Joint continuous update.} We then re-enable gradients for the encoder and perform a second, fully differentiable update on the same minibatch, feeding the continuous probabilities $p$ to the predictor and
    backpropagating through both encoder and predictor under the full objective (prediction plus regularizers). This step nudges the encoder to produce continuous codes that, after binarization, remain easy to model for a predictor already specialized to discrete inputs.
\end{enumerate}

Although this procedure doubles the number of optimization steps per minibatch, notice that the computational overhead remains modest: the heavier model is the encoder and it can be evaluated only once per batch.
%The encoder is evaluated only once per minibatch, so the extra cost is due mainly to the additional predictor update. 
%In the ablation of Section~\ref{subsec:ablation-training}, we compare this two-step scheme against two single-step procedures: a fully differentiable one with predictor on $p$, and a straight-through variant. The proposed two-step procedure yields equal or better roll-out accuracy.
Moreover, as shown in Section~\ref{subsec:ablation-training}, the proposed procedure achieves equal or superior accuracy than a straight-through estimator or a fully differentiable training with soft values.

\section{Experiments}\label{sec:experiments}
We evaluate our framework on two combinatorially challenging domains:  \emph{MNIST 8-puzzle} and \emph{IceSlider}. 
%a variation of the sliding 8-puzzle rendered with MNIST digits, and the IceSlider benchmark. 
In both settings, we assess the ability of the models to learn a compact, discrete latent representation that captures the environment's known symbolic states and dynamics without supervision. 
%the \emph{MNIST 8-puzzle}, used throughout the paper, and \emph{IceSlider}, introduced by Bagatella et al.~\cite{bagatella2021planning} and adopted as a benchmark in DeepCubeAI~\cite{agostinelli2024deepcubeai}. 
Code to reproduce all experiments is available at \href{https://github.com/dbizzaro/DWMR}{https://github.com/dbizzaro/DWMR}.
\subsection{Benchmarks}\label{sec:dataset}
%We use the $8$-puzzle on a $3\times3$ grid with tiles $\{1,\dots,8\}$ and a blank tile $0$.
\subsubsection{MNIST 8-puzzle.} %We first consider the 8-puzzle benchmark rendered with MNIST digits.
Each environment state is a configuration of a $3{\times}3$ sliding puzzle with tiles $\{1,\dots,8\}$ and a blank tile $0$. Actions are $\{\texttt{up,down,left,right}\}$, and move the blank cell; invalid moves (that would push the blank cell outside of the grid) are rejected. States are rendered as single-channel $88{\times}88$ images by placing MNIST patches on a $3{\times}3$ grid with 1-pixel gutters; the digit for each tile is re-sampled at every timestep, while the blank is rendered as a black square (see Fig.~\ref{fig:model}). Optionally, to test the  robustness to perceptual noise, we add i.i.d.\ Gaussian pixel noise $\mathcal{N}(0,0.5)$ and clip intensities to $[0,1]$. Train/validation/test splits are generated as long random-walk trajectories from random solvable initial states, with distinct seeds across splits (to test generalization to new trajectories and configurations) and disjoint
MNIST sources across splits (to test perceptual generalization to unseen digit instances). We use $30{,}000$ transitions for training and $6{,}000$ for validation and for test.

\subsubsection{IceSlider.}
\begin{wrapfigure}{r}{0.48\textwidth}
  \centering
  \vspace{-0.5\baselineskip} % optional: tighten vertical spacing
%\begin{figure}[ht]
  %\centering
  \includegraphics[width=\linewidth]{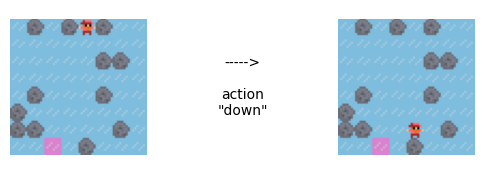}
  \caption{
    Example transition in IceSlider. 
    %Each board is a $3 \times 3$ sliding puzzle with a blank tile and tiles numbered from $1$ to $8$ rendered by sampling a fresh MNIST digit. Each action induces a local state transition by swapping the blank with an adjacent tile.
  }
  \label{fig:iceslider}
  \vspace{-1.6\baselineskip} % optional
%\end{figure}
\end{wrapfigure}

IceSlider is an $8{\times}8$ grid puzzle introduced in \cite{bagatella2021planning} and adopted as a benchmark for DeepCubeAI~\cite{agostinelli2024deepcubeai}. Observations are rendered as $64{\times64}$ RGB images depicting the $8{\times}8$ board, using fixed patches for the agent, the goal cell, the rocks (obstacles), and the ice (free cells). When an action $\{\texttt{up,down,left,right}\}$ is taken, the agent moves in the chosen direction until it is stopped by either a rock or the boundary of the grid (see Fig.~\ref{fig:iceslider}). 
We generate an offline dataset from random interaction episodes consisting of $20$ actions, and follow previous work \cite{bagatella2021planning,agostinelli2024deepcubeai} but with a tenth of the repetitions: we collect $40{,}000$ overall transitions for training and $10{,}000$ for validation and for test. Similarly to the 8-puzzle, we report results both in a clean setting and under Gaussian $\mathcal{N}(0,0.5)$ pixel noise.

\subsection{Experimental Setup}
%\subsection{Evaluation Metrics} 
%\subsection{Implementation Details}

\subsubsection{Architectures.} For the \emph{MNIST 8-puzzle}, the encoder $enc_\phi$ is a 5-layer CNN with $3{\times}3$ convolutions, GroupNorm and ReLU, and with $2{\times}2$ average pooling after the first three layers. %; thus mapping $88{\times}88$ images to a feature grid of size $11{\times}11$. 
The final grid is flattened and passed through a two-layer MLP (hidden size $96$) to produce $K=64$ sigmoid outputs $p\in[0,1]^K$, thresholded at $0.5$ into bits $b\in\{0,1\}^K$. 
The predictor $pred_\psi$ is a three-layer MLP %that takes the concatenation of $p$ (or $b$) and the one-hot action $a\in\{0,1\}^4$, 
with hidden layer of size $128$, ReLU activation, and sigmoid outputs. 
%outputs Bernoulli parameters for the next latent state. 

For \emph{IceSlider}, we adopt instead the grid-aligned architecture used in DeepCubeAI \cite{agostinelli2024deepcubeai}. Concretely, the encoder $enc_\phi$ is made by a convolutional layer with kernel size $4$, stride $4$, ReLU activation, and $32$ channels, followed by another convolutional layer with kernel size $2$, stride $2$, sigmoid output and $3$ channels. The predictor $pred_\psi$ concatenates on the channel dimension the $3{\times}8{\times}8$ image encodings (indeed, $K{=}192$) with the one-hot actions, and pass everything to four residual blocks with kernel size $3$, batch normalization, and ReLU activations.
%, and (ii) the predictor is a \emph{convolutional residual}
%transition model that is action-conditioned and predicts the next latent tensor.
%This inductive bias differs from 8-game, where the latent is a flat bit-vector and the predictor is an MLP.

In both benchmarks, when a decoder $dec_\eta$ is used (as in all baselines but DWMR), we employ a convolutional decoder mirroring the encoder architecture.

\subsubsection{Evaluation Metrics.}
We evaluate representation quality with supervised \emph{linear probes} fitted after world-model training: we freeze the encoder and fit a lightweight readout that tries to reconstruct the underlying symbolic grid state.
%We evaluate all models using \emph{linear probes} that try to reconstruct the full symbolic board configuration from the learned latent representation.
%So, after training a world model, we freeze its encoder and attach a small supervised readout that operates on the encoder output for each frame $x$. 
%Concretely, this readout is an affine (fully connected) layer that maps each Boolean latent code $b$ to $9$ independent softmax classifiers: one for each cell of the $3 \times 3$ grid, each  predicting a digit in $\{0, \dots, 8\}$. 
For the \emph{MNIST 8-puzzle}, the probe is a single affine layer mapping the Boolean latent vector $b$ to $9$ independent softmax classifiers (one per board cell), each predicting a value in $\{0,\dots,8\}$. %And we report \emph{mean per-cell accuracy} (averaging across the 9 cells).
Instead, for \emph{IceSLider}, we use a single-layer convolutional probe with kernel size $1$, to predict per-cell classes $\{\texttt{ice,rock,agent,goal}\}$ over the $8{\times}8$ grid. In both cases, the metric is \emph{mean per-cell f1-score}: the f1-score for each cell classification, averaged across the $9$ (\emph{8-puzzle}) or $64$ (\emph{IceSlider}) cells.
%We tried also deeper probes, but the results were not substantially different from the linear ones.

The probes are trained using the training split to fit the parameters, and the test split to report final performance. We use cross-entropy loss, train for $15$ epochs with Adam optimizer, and fix the learning rate to $0.01$ with a weight decay of $0.001$. Given the trained probes, we report the performance metric in two settings:
\begin{itemize}
    \item \textbf{Encoding (Enc.)}: the probe is applied directly to the Boolean latent codes at each timestep.
    \item \textbf{Imagination (Im.)}: the probe is applied after rolling-out the discrete latent dynamics in imagination for one step, using the predictor network.
\end{itemize}
%Both the world models and the probes are optimized with the Adam optimizer.

\subsubsection{Optimization and tuning.} 
All models are trained with Adam optimizer and batch size $N=256$, for a fixed budget of $40$ epochs for MNIST 8-puzzle, and $20$ epochs for IceSlider. For each benchmark, the compared models share the same architecture for the encoder and for the predictor: differences are due only to the presence of the decoder and to the loss terms. For every model family, benchmark and noise setting, we perform a hyperparameter search using the probe imagination results on the validation set. Final test results are computed using the model instance corresponding to the chosen hyperparameters, averaging the results over $10$ runs. 
%We allow distinct learning rates for the encoder and predictor. 
In all experiments we fix $\lambda_{\text{pred}}=1$, while tuning the coefficients of the remaining loss terms (i.e., when applicable, $\lambda_{\text{var}}, \lambda_{\text{cor}}, \lambda_{\text{cos}}, \lambda_{\text{loc}}$, $\lambda_{\text{rec}}$ and $\lambda_{\text{KL}}$). We also tune the learning rate, the locality window bounds $L$ and $U$, and the EMA decay coefficient $\tau$ of the target encoder. Finally, we also apply exponential scheduling of such hyperparameters, tuning also the scheduling coefficients.

\subsection{Model Comparison}

We compare DWMR to some variants and to discrete autoencoding baselines:
\begin{itemize}
\item\textbf{AE.}
This baseline augments the world model with a convolutional decoder $\mathrm{dec}_\eta$ trained to reconstruct pixels from encoder probabilities $p$. The training loss is 
$
\mathcal{L}_{AE}\coloneqq \mathcal{L}_{\text{pred}} + \lambda_{\text{rec}} \mathcal{L}_{\text{rec}}$, with $\mathcal{L}_{\text{rec}} \coloneqq \mathrm{MSE}\bigl(\mathrm{dec}_\eta(p), x\bigr)$.

\item\textbf{$\beta$-VAE.} The (Binary Concrete) $\beta$-VAE baseline uses the same decoder architecture as AE, but its inputs $p$ are formed adding logistic noise to the logits (just before the encoder final $sigmoid$ activation) \cite{asai2022classical}. In addition to prediction and reconstruction, it also adds a KL penalty to a $Bernoulli(0.5)$ prior,
\[
\mathcal{L}_{\text{KL}}
%= \frac{1}{K}\sum_{k=1}^K
%\mathrm{KL}\!\bigl(\mathrm{Bern}(p_{k}) \,\|\, \mathrm{Bern}(0.5)\bigr)
\coloneqq \mathbb{E}_n \frac{1}{K}\sum_{k=1}^K \Bigl[
p_{nk}\log\frac{p_{nk}}{0.5} + (1-p_{nk})\log\frac{1-p_{nk}}{0.5}
\Bigr],
\]
which nudges each bit towards the ``undecided'' value $p_{nk}=0.5$. So the overall loss is $
\mathcal{L}_{\beta\text{-VAE}} \coloneqq \mathcal{L}_{\text{pred}} + \lambda_{\text{rec}} \mathcal{L}_{\text{rec}} + \lambda_{\text{KL}}\mathcal{L}_{\text{KL}}$.
%again without our explicit Boolean regularizers.
%The VAE baseline further adds a KL term between the approximate posterior and a factorized prior on the latent units, weighted by its own coefficient. Its loss combines prediction, reconstruction, and KL.

\item \textbf{DeepCubeAI.} This is a discrete autoencoding world model with the loss function used in DeepCubeAI~\cite{agostinelli2024deepcubeai}. In particular, let $p'$ be the soft encoding of next step observation $x'$, $\hat p'$ be the output of the predictor, $r(\cdot)$ be a rounding function with a straight-through estimator, and $\mathrm{sg}(\cdot)$ be stop-gradient.
Then, encoder, decoder and predictor are trained jointly with $
\mathcal{L}_{\text{DeepCubeAI}}
:= \mathcal{L}_{\text{pred}'}
+ \lambda_{\text{rec}} \mathcal{L}_{\text{rec}'},
$
where
\[
\mathcal{L}_{\text{rec}'}
\coloneqq \tfrac{1}{2}\mathrm{MSE}(\mathrm{dec}_\eta(p), x)
+ \tfrac{1}{2}\mathrm{MSE}(\mathrm{dec}_\eta(p'), x'),
\]
and
\[
\mathcal{L}_{\text{pred}'}
\coloneqq \tfrac{1}{2}\mathrm{MSE}(r(p'),\mathrm{sg}(r(\hat p')))
+ \tfrac{1}{2}\mathrm{MSE}(\hat p',\mathrm{sg}(r(p'))).
\]
%Here $p'$ is the soft encoding of $x'$, $\hat p'$ is the next-step prediction, $r(\cdot)$ denotes rounding with a straight-through estimator, and $\mathrm{sg}(\cdot)$ is stop-gradient.

\item \textbf{DWMR.}
Our Discrete World Model via Regularization is trained only with the predictive objective and the four regularizers of Eq.~\eqref{eq:loss}.

\item \textbf{DWMR{+}AE/$\beta$-VAE.}
In these variants, we keep the full $\mathcal{L}_{\text{DWMR}}$ loss of Eq.~\eqref{eq:loss} and \emph{add} a decoder (deterministic or variational) as an auxiliary head. The objectives are therefore $\mathcal{L}_{\text{DWMR}}+\lambda_{\text{rec}}\mathcal{L}_{\text{rec}}$ and $\mathcal{L}_{\text{DWMR}}+\lambda_{\text{rec}}\mathcal{L}_{\text{rec}} + \lambda_{\text{KL}}\mathcal{L}_{\text{KL}}$ respectively.
\begin{comment}
Its world model is trained with a reconstruction loss
\[
\mathcal{L}_{\text{rec}} \coloneqq 
= \frac{1}{2N}\sum_{i=1}^N
\bigl\lVert s_i - \hat{s}_i\bigr\rVert_2^2
+ \frac{1}{2N}\sum_{i=1}^N
\bigl\lVert s'_i - \hat{s}'_i\bigr\rVert_2^2,
\]
on current and next states, and a latent model loss
\[
L_m(\theta)
= \frac{1}{2N}\sum_{i=1}^N
\Bigl(
\bigl\lVert r(\tilde{s}'_i) - r(\hat{\tilde{s}}'_i)\bigr\rVert_2^2
+ \bigl\lVert r(\tilde{s}'_i)^{\text{det}} - \hat{\tilde{s}}'_i\bigr\rVert_2^2
\Bigr),
\]
which symmetrizes the mean-squared error between rounded next-step latents from the encoder and from the transition model (with straight-through rounding and detached targets).\footnote{We follow the notation of~\cite{agostinelli2025learning}, where tildes denote latent states and $r(\cdot)$ is the rounding operator implemented with a straight-through estimator.} These are combined as
\[
L(\theta) = (1-\omega)\,L_r(\theta) + \omega\,L_m(\theta),
\]
with a schedule on $\omega$ that gradually balances reconstruction and model consistency. In the full DeepCubeAI system, the learned model is then used to generate data for a DQN heuristic trained with a squared TD loss but in our comparison we only use the world-model component.
\end{comment}
\end{itemize}

\subsubsection{Results.} 
Table~\ref{tab:models} reports per-cell probe f1-scores for reconstructing the board configuration from encoder boolean states (\emph{Enc.} columns) and from one-step imagination roll-outs~(\emph{Im.} columns). %, both with clean images and with Gaussian noise addition. 

\begin{table}[ht]
    \centering
    \caption{Mean per-cell f1-scores, for reconstructing the board configurations from encoder states (\emph{Enc.}) and from one-step roll-outs in imagination (\emph{Im.}), using linear probes. 
    We report the mean and standard deviation over $10$ different runs, for each model and benchmark.} %All models are trained with our two steps pro and hyperparameters scheduling.}
    \label{tab:models}
    
\begin{tabular}{l*{2}{ll}*{2}{ll}}
    \toprule
    \multicolumn{1}{l}{Models} &
    \multicolumn{4}{c}{MNIST 8-puzzle} &
    \multicolumn{4}{c}{IceSlider}
    \\
    \cmidrule(lr){2-5}\cmidrule(lr){6-9}
    & \multicolumn{2}{c}{Clean} &
      \multicolumn{2}{c}{Noisy} &
      \multicolumn{2}{c}{Clean} &
      \multicolumn{2}{c}{Noisy}
    \\
    \cmidrule(lr){2-3}\cmidrule(lr){4-5}\cmidrule(lr){6-7}\cmidrule(lr){8-9}
    & ~Enc. & ~Im. & ~Enc. & ~Im. & ~Enc. & ~Im. & ~Enc. & ~Im.
    \\
    \midrule

    AE & 41 \scriptsize{±26} & 36 \scriptsize{±22} 
    & 50 \scriptsize{±10} & 44 \scriptsize{±8} 
    & 96 \scriptsize{±11} & 83 \scriptsize{±11} 
    & 83 \scriptsize{±8} & 84 \scriptsize{±9} 
    \\

    $\beta$-VAE & 74 \scriptsize{±2} & 61 \scriptsize{±2} 
    & 70 \scriptsize{±1}  & 58 \scriptsize{±1}  
    & 93 \scriptsize{±10} & 62 \scriptsize{±13} 
    & 70 \scriptsize{±14} & 58 \scriptsize{±4} 
    \\

    DeepcubeAI & 79 \scriptsize{±3} & 77 \scriptsize{±4} 
    & 74 \scriptsize{±2} &  73 \scriptsize{±2} 
    & 92 \scriptsize{±14} & 85 \scriptsize{±14} 
    & 81 \scriptsize{±14} & 80 \scriptsize{±13} 
    \\

    \midrule

    DWMR & 91 \scriptsize{±1} & 84 \scriptsize{±2}
    & 90 \scriptsize{±2} & 80 \scriptsize{±3} 
    & \textbf{99} \scriptsize{±2} & \textbf{95} \scriptsize{±2} 
    & 81 \scriptsize{±6} & 85 \scriptsize{±5} 
    \\

    DWMR{+}AE & 94 \scriptsize{±2} & 86 \scriptsize{±3} 
    & \textbf{98} \scriptsize{±1} & \textbf{90} \scriptsize{±3} 
    & \textbf{99} \scriptsize{±2} & 91 \scriptsize{±10} 
    & \textbf{86} \scriptsize{±6} & \textbf{88} \scriptsize{±6} 
    \\

    DWMR{+}$\beta$-VAE & \textbf{96} \scriptsize{±5} & \textbf{91} \scriptsize{±4} 
    & 92 \scriptsize{±2} & 82 \scriptsize{±1} 
    & \textbf{99} \scriptsize{±2} & 90 \scriptsize{±4} 
    & 73 \scriptsize{±12} & 70 \scriptsize{±9} 
    \\

    \bottomrule
\end{tabular}
\end{table}

Across all settings, DWMR achieves strong and stable encoding and roll-out performances, despite being trained without any pixel-level reconstruction signal. 
In contrast, baselines that rely primarily on generative reconstruction (AE, $\beta$-VAE, and DeepCubeAI) consistently underperform.
Among these, DeepCubeAI improves over AE and $\beta$-VAE, but still falls short of DWMR.
%Generative baselines that rely only on reconstruction (AE, $\beta$-VAE, and DeepCubeAI) consistently underperform our reconstruction-free model, indicating that pixel reconstruction alone does not enforce a propositional, dynamically useful code. 
%Adding a decoder on top of our losses (DWMR + AE / DWMR + $\beta$-VAE) yields systematic but relatively modest improvements over DWMR, compared to the much larger gap between DWMR and the generative-only baselines. 
%Overall, most of the useful structure in the Boolean latents comes from the dynamics-driven objective and explicit regularizers, while generative heads act mainly as auxiliary signals.
%Adding an auxiliary decoder on top of DWMR, as in DWMR+$\beta$-VAE and especially in DWMR+AE, yields further gains, but the improvements are comparatively modest relative to the large gap between our reconstruction-free model and the regularization-free baselines.
%Overall, these results support our claim that most of the useful structure in Boolean latents is induced by a dynamics-driven objective paired with explicit factorization and locality regularization, while generative heads act mainly as auxiliary signals rather than the primary mechanism preventing collapse. 
%Overall, these results indicate that our Boolean-specific regularizers are sufficient to learn a non-collapsed state space, and that generative components provide only secondary benefits in this setting.
Overall, the results support the conclusion that the proposed Boolean-specific regularizers are sufficient to learn a non-collapsed state space and transition function, even outperforming decoder-only alternatives. Finally, augmenting DWMR with an auxiliary decoder (as in DWMR+AE) can further improve performance, suggesting that reconstruction can provide complementary benefits once the latent space is already well-shaped by the regularization.
%Overall, these results indicate that our Boolean-specific regularizers are sufficient to learn a non-collapsed state space and transition function, even outperforming decoder-only alternatives. %In th adding generative components provide significant but secondary benefits in this setting.
%Moreover, adding a decoder on top of our regularized objective (as in DWMR+AE) can further improve performance, yielding additional gains beyond an already strong reconstruction-free baseline.

\section{Ablations}
In this section, we ablate key training and objective components, restricting the analysis to the \emph{MNIST 8-puzzle} benchmark, and reporting per-cell accuracy (rather than per-cell f1-score, which was motivated by IceSlider class imbalance).

\subsection{Comparison of Training Variants}\label{subsec:ablation-training}

To study the trade-off between discrete and differentiable optimization in the prediction loss, we considered three training variants that differ only in how the predictor receives its input and how gradients are propagated through the encoder. %As always, let $p$ denote the encoder probabilities, $b$ the corresponding binarized codes, and $\hat{p}'$ the predicted probabilities for the target bits $b'$. 
Contextually, we also validate the use of simple exponential schedules for the hyperparameters. With them, at the end of each epoch, the learning rates, the EMA coefficient $\tau$, and the loss weights are each multiplied by a different fixed factor in $[0.9,1.1]$ selected during hyperparameter search.
%We try also applying simple exponential schedules to a set of scalar hyperparameters. At the end of each epoch, the learning rate, the EMA coefficient $\tau$, and the loss weights are each multiplied by a different fixed factor in $[0.9,1.1]$ selected during hyperparameter search. This gradually shifts the relative strength of the regularizers over training, without introducing additional losses or a hand-crafted curriculum.
The three different training regimes are the following: (i) a \emph{fully differentiable} scheme where the predictor always sees soft probabilities, (ii) a \emph{straight-through} scheme where the predictor sees hard bits but gradients flow as if probabilities had been used, and (iii) the proposed \emph{two-steps} combination, where we first update the predictor on hard bits and then update encoder and predictor jointly with a straight-through estimator. %All other architectural and loss components are identical.

\begin{table}[ht]
    \centering
    \caption{Mean per-cell accuracies, for reconstructing the board configurations from encoder states and from one-step roll-outs in imagination, using linear probes. 
    We report the mean and standard deviation over $10$ different runs on the \emph{MNIST 8-puzzle} benchmark.}
    \label{tab:settings}
    
    \begin{tabular}{lc*{2}{cc}}
    \toprule
    \multicolumn{1}{l}{Model} &
    \multicolumn{1}{l}{Scheduling} &
    \multicolumn{2}{c}{Clean} &
    \multicolumn{2}{c}{Noisy} %&
    %\multicolumn{2}{c}{Perlin noise}
    \\
    \cmidrule(lr){3-4}\cmidrule(lr){5-6}%\cmidrule(lr){7-8}
    & &  Encoding & Imagination & Encoding & Imagination %&  Encoding & Roll-out 
    \\
    \midrule

    \multirow{2}[1]{*}{Fully differentiable } & yes 
    & 95 \scriptsize{±2} & 80 \scriptsize{±2} 
    & 92 \scriptsize{±2} & 74 \scriptsize{±2} 
    %& \textbf{90} \scriptsize{±2} & 75 \scriptsize{±1}
    \\
    \cline{3-6}
    & no 
    & 93 \scriptsize{±5} & 76 \scriptsize{±3} 
    & 85 \scriptsize{±19} & 75 \scriptsize{±16} 
    %& 78 \scriptsize{±9} & 69 \scriptsize{±8} 
    \\
    \midrule
    
    \multirow{2}[1]{*}{Straight-through} & yes 
    & \textbf{96} \scriptsize{±1} & 85 \scriptsize{±1} 
    & 88 \scriptsize{±15} & 81 \scriptsize{±13} 
    %& \textbf{90} \scriptsize{±2} & 78 \scriptsize{±2} 
    \\
    \cline{3-6}
    & no 
    & \textbf{96} \scriptsize{±4} & 80 \scriptsize{±2} 
    & 86 \scriptsize{±4} & 77 \scriptsize{±4} 
    %& 80 \scriptsize{±26} & 66 \scriptsize{±20} 
    \\
    \midrule
    
    \multirow{2}[1]{*}{Two steps} & yes 
    & 93 \scriptsize{±6} & \textbf{86} \scriptsize{±5} 
    & \textbf{95} \scriptsize{±2} & \textbf{83} \scriptsize{±3} 
    %& \textbf{90} \scriptsize{±8} & \textbf{82} \scriptsize{±1} 
    \\
    \cline{3-6}
    & no 
    & 92 \scriptsize{±4} & 84 \scriptsize{±4} 
    & 91 \scriptsize{±3} & 81 \scriptsize{±2} 
    %& \textbf{90} \scriptsize{±8} & 80 \scriptsize{±7} 
    \\    
    %\midrule
    
    %Two steps - MSE loss & yes & \textbf{96} \scriptsize{±1} & 85 \scriptsize{±1} & 92 \scriptsize{±2} & 77 \scriptsize{±2} & \textbf{92} \scriptsize{±3} & 77 \scriptsize{±3} \\
    
    \bottomrule
    \end{tabular}
\end{table}

Empirically, the \emph{two-steps} scheme is competitive or slightly better than the others in terms of encoding and roll-out accuracy,  in particular under Gaussian noise. Scheduling the main hyperparameters further improves robustness: across all three variants, the scheduled runs are at least as good as their fixed counterparts and often yield better results. %For the \emph{two-steps} variant in particular, scheduling leads to small but consistent gains and lower variance across runs. 
Thus, we used the \emph{two-steps} training scheme with scheduled hyperparameters as our default configuration.

\subsection{Ablation on Model Parts}

Table~\ref{tab:ablation} reports the results of an ablation study where we start from the best hyperparameters of our reconstruction-free DWMR model, and remove individual terms from the objective (or, in \emph{No~EMA}, let the parameters $\phi'$ be equal to $\phi$, instead of being an exponential moving average of $\phi$).  
%Across all settings, dropping the variance regularizer $\mathcal{L}_{\text{var}}$ completely destroys performance,  %{\color{red} ... (TODO)}
%Removing the decorrelation term \(L_{\mathrm{cor}}\) also leads to a large and systematic degradation, confirming that discouraging co-adapted bits is important for a useful code. The coskewness penalty \(L_{\mathrm{cos}}\) has a milder but still visible effect, especially on roll-out accuracy and robustness to noise.

%\setlength{\columnsep}{1.5em}
%\begin{wraptable}{r}{0.60 \textwidth}
\begin{table}[ht]
%\vspace{-1.6\baselineskip}  
    \centering
    \caption{Ablation study starting from the DWMR hyperparameters. We report the mean and standard deviation over 10 different runs on the \emph{MNIST 8-puzzle} benchmark of the mean per-cell accuracies.}
    \label{tab:ablation}
%\vspace{-0.5\baselineskip}  
\begin{tabular}{l*{3}{ll}}
    \toprule
    \multicolumn{1}{l}{Model} &
    \multicolumn{2}{c}{Clean} &
    \multicolumn{2}{c}{Noisy} %&
    %\multicolumn{2}{c}{Perlin noise}
    \\
    \cmidrule(lr){2-3}\cmidrule(lr){4-5}%\cmidrule(lr){6-7}
    &  ~Enc. & ~Im. & ~Enc. & ~Im. %&  Encoding & Roll-out 
    \\
    \midrule

    DWMR & 93 \scriptsize{±6} & 86 \scriptsize{±5} 
    & 95 \scriptsize{±2} & 83 \scriptsize{±3} 
    %& 90 \scriptsize{±8} & 82 \scriptsize{±1} 
    \\

    No $\mathcal{L}_{\text{var}}$ & 12 \scriptsize{±0} & 12 \scriptsize{±0} & 11 \scriptsize{±1} & 11 \scriptsize{±1} 
    %& 12 \scriptsize{±0} & 12 \scriptsize{±0} 
    \\

    No $\mathcal{L}_{\text{cor}}$ & 42 \scriptsize{±8} & 39 \scriptsize{±6} & 39 \scriptsize{±4} & 35 \scriptsize{±3} 
    %& 50 \scriptsize{±4} & 45 \scriptsize{±3} 
    \\

    No $\mathcal{L}_{\text{cos}}$ & 56 \scriptsize{±13} & 53 \scriptsize{±11} 
    & 67 \scriptsize{±14} & 60 \scriptsize{±11} 
    %& 51 \scriptsize{±13} & 48 \scriptsize{±11} 
    \\

    No $\mathcal{L}_{\text{loc}}$ & 89 \scriptsize{±15} & 82 \scriptsize{±14}
    & 38 \scriptsize{±24} & 33 \scriptsize{±23} 
    %& 47 \scriptsize{±31} & 43 \scriptsize{±30} 
    \\

    %No $\mathcal{L}_{\text{pred}}$ (on $2^{nd}$ step) & 71 \scriptsize{±6} & 61 \scriptsize{±5} & 
    %77 \scriptsize{±2} & 58 \scriptsize{±2} & 
    %81 \scriptsize{±4} & 67 \scriptsize{±6} \\

    No EMA & 87 \scriptsize{±15} & 81 \scriptsize{±13} 
    & 93 \scriptsize{±2} & 79 \scriptsize{±4}
    %& 77 \scriptsize{±25} & 67 \scriptsize{±28} 
    \\
    \bottomrule
    \end{tabular}
\end{table}
%\vspace{-1.2\baselineskip}
%\end{wraptable}

This ablation study confirms that each component of the proposed loss function plays a distinct role in shaping the latent space:
\begin{itemize}
    \item \textbf{Prevention of latent collapse.} The variance regularizer $\mathcal{L}_{var}$ is the most critical component: its removal leads to a complete collapse of the latent space, with accuracy dropping to random guessing levels across all settings. This confirms that without explicitly enforcing variability, the Boolean code degenerates into a constant, non-informative state.

    \item \textbf{Bit independence.} The correlation regularizer $\mathcal{L}_{cor}$ and in lesser terms the coskewness regularizer $\mathcal{L}_{cos}$ are essential for a good representation. Removing any of them significantly degrades performance, indicating that penalizing pairwise and higher-order dependencies is necessary to force the bits to capture distinct factors of variation. %The coskewness regularizer $\mathcal{L}_{cos}$ complements this by suppressing higher-order dependencies, providing a consistent improvement in representation quality.

    \item \textbf{Robustness.} The effect of the locality regularizer $\mathcal{L}_{\text{loc}}$ appears to depend on the chosen hyperparameters: in the clean setting, dropping $\mathcal{L}_{\text{loc}}$ degraded the performance of only one out of ten runs, while in the noisy setting the degradation was much more pronounced. 
    %The impact of the locality regularizer $\mathcal{L}_{loc}$ is more nuanced and heavily dependent on the chosen hyperparameters (specifically the target window $[L, U]$). In clean environments, removing $\mathcal{L}_{loc}$ results in accuracy comparable to the base model, suggesting that it is not strictly necessary for pure prediction when the signal is strong. However, even when it has little effect on final accuracy, this term enforces a structural prior that biases the model toward a less distributed representation. By constraining transitions to flip only a sparse subset of bits, the model aligns closer to the combinatorial nature of the environment, potentially aiding interpretability and robustness in noisier regimes.
    %\item \textbf{Robustness.} Finally, 
    Similarly, also the use of EMA on the target encoder proves important for robustness. %more than for peak performance, at least: removing the EMA mechanism lead one run on the \emph{Clean} setting and two on the \emph{Perlin noise} setting to remain stuck at chance level or slightly above, lowering the mean performance and increasing a lot the variance.

    %\item {\color{red} Add (already calculated) ablation term given by removing $\mathcal{L}_{pred}$ on the training scheme's second step (so with predictor trained alone, without providing signal to the encoder)? Add ablation term without hyperparameters scheduling?}
\end{itemize}

%Supplementary Material reports analogous ablations starting from DWMR+AE and DWMR+$\beta$-VAE, which substantially confirm the discussion above. 

%The main difference is only that in most settings dropping the locality regularizer does not harm the results, although one may still want locality to ensure a less distributed and therefore more interpretable representation.

%In summary, while $\mathcal{L}_{pred}$ drives the dynamics learning, the regularizers are essential for shaping the Boolean space: $\mathcal{L}_{var}$ ensures the code carries information, $\mathcal{L}_{cor/cos}$ enforces independence, and $\mathcal{L}_{loc}$ imposes a sparse, logical structure on the transitions.

\section{Limitations and Future Directions}
While our results are encouraging, important challenges remain that open up several natural avenues for future work. First, our experiments do not address \emph{partial observability}, for which it may be necessary to incorporate some memory (e.g., with a recurrent architecture). More explicit treatment of \emph{stochastic environments} and/or \emph{continuous actions} are also important for broader applicability.  In general, evaluating on \emph{more domains}, with different dynamics, would help clarify the scope and robustness of the approach. Second, a natural direction is to integrate our model with either MBRL policy learning or with planning/heuristic search to test the \emph{performance on concrete goal-driven tasks}. Finally, given the locality prior, it would be interesting to study the \emph{interpretability of the learned transition functions}, and whether locality yields more modular, human-readable update rules.

\section{Conclusions}
In this work, we introduced a novel framework for learning world models with discrete Boolean latent states, possibly removing the reliance on generative pixel-level reconstruction. Central to our approach is the design of a specialized regularization objective that explicitly shapes the latent space: we go beyond simple variance/covariance constraints by introducing \emph{coskewness} regularization to enforce higher-order bit independence, and a \emph{locality} regularizer that imposes a structural prior for sparse, action-induced state changes. Furthermore, we proposed a novel training scheme with \emph{two-step updates} ensuring that the predictor remains robust to the binary inputs it encounters at test time, while using backpropagation for training.
Our experiments on two benchmarks requiring strong generalization to novel board configurations demonstrate that this regularization approach yields superior state representations compared to autoencoding baselines, %These results were achieved in a challenging regime requiring strong generalization, : the test set consisted of virtually entirely \emph{novel board configurations} and unseen digit renderings, with negligible overlap to the training trajectories. 
and ablations indicate that each design choice contributes to the final representation quality.
Overall, this confirms that our tailored Boolean regularizers successfully drive the world models to capture the underlying symbolic structure of the environments. Finally, while DWMR removes the need for pixel-level reconstruction, it does not preclude it: indeed, pairing DWMR with an auxiliary reconstruction decoder yield additional gains. 

\printbibliography

@article{hafner2019dreamer,
  title={Dream to control: Learning behaviors by latent imagination},
  author={Hafner, Danijar and Lillicrap, Timothy and Ba, Jimmy and Norouzi, Mohammad},
  journal={arXiv preprint arXiv:1912.01603},
  year={2019}
}

@article{hafner2020dreamerv2,
  title={Mastering atari with discrete world models},
  author={Hafner, Danijar and Lillicrap, Timothy and Norouzi, Mohammad and Ba, Jimmy},
  journal={arXiv preprint arXiv:2010.02193},
  year={2020}
}

@article{hafner2025dreamerv3,
  title={Mastering diverse control tasks through world models},
  author={Hafner, Danijar and Pasukonis, Jurgis and Ba, Jimmy and Lillicrap, Timothy},
  journal={Nature},
  year={2025},
}

@inproceedings{okada2020dreaming,
  title={Dreaming: Model-based reinforcement learning by latent imagination without reconstruction},
  author={Okada, Masashi and Taniguchi, Tadahiro},
  booktitle={2021 ieee international conference on robotics and automation (icra)},
  pages={4209--4215},
  year={2021},
  organization={IEEE}
}

@inproceedings{deng2022dreamerpro,
  title={Dreamerpro: Reconstruction-free model-based reinforcement learning with prototypical representations},
  author={Deng, Fei and Jang, Ingook and Ahn, Sungjin},
  booktitle={International conference on machine learning},
  year={2022},
}

@article{burchi2024mudreamer,
  title={MuDreamer: Learning Predictive World Models without Reconstruction},
  author={Burchi, Maxime and Timofte, Radu},
  journal={arXiv preprint arXiv:2405.15083},
  year={2024}
}

@misc{hansen2024tdmpc2,
	title={TD-MPC2: Scalable, Robust World Models for Continuous Control}, 
	author={Nicklas Hansen and Hao Su and Xiaolong Wang},
	booktitle={International Conference on Learning Representations (ICLR)},
	year={2024}
}

@inproceedings{assran2023ijepa,
  title={Self-supervised learning from images with a joint-embedding predictive architecture},
  author={Assran, Mahmoud and Duval, Quentin and Misra, Ishan and Bojanowski, Piotr and Vincent, Pascal and Rabbat, Michael and LeCun, Yann and Ballas, Nicolas},
  booktitle={Proceedings of the IEEE/CVF Conference on Computer Vision and Pattern Recognition},
  year={2023}
}

@article{bardes2024vjepa,
  title={Revisiting Feature Prediction for Learning Visual Representations from Video},
  author={Bardes, Adrien and Garrido, Quentin and Ponce, Jean and Rabbat, Michael and LeCun, Yann and Assran, Mahmoud and Ballas, Nicolas},
  journal={arXiv:2404.08471},
  year={2024}
}

@article{zhou2024dinowm,
    title={DINO-WM: World Models on Pre-trained Visual Features enable Zero-shot Planning}, 
    author={Gaoyue Zhou and Hengkai Pan and Yann LeCun and Lerrel Pinto},
    journal={arXiv preprint arXiv:2411.04983},
    year={2024}
}

@inproceedings{bardes2022vicreg,
  author  = {Adrien Bardes and Jean Ponce and Yann LeCun},
  title   = {VICReg: Variance-Invariance-Covariance Regularization For Self-Supervised Learning},
  booktitle = {ICLR},
  year    = {2022}
}

@article{asai2022classical,
  title={Classical planning in deep latent space},
  author={Asai, Masataro and Kajino, Hiroshi and Fukunaga, Alex and Muise, Christian},
  journal={Journal of Artificial Intelligence Research},
  volume={74},
  pages={1599--1686},
  year={2022}
}

@inproceedings{dittadi2021atari,
  title={Planning from pixels in atari with learned symbolic representations},
  author={Dittadi, Andrea and Drachmann, Frederik K and Bolander, Thomas},
  booktitle={Proceedings of the AAAI Conference on Artificial Intelligence},
  year={2021}
}

@article{lecun2022path,
  title={A Path Towards Autonomous Machine Intelligence},
  author={LeCun, Yann},
  journal={arXiv preprint arXiv:2201.05966},
  year={2022}
}

@article{agostinelli2024deepcubeai,
  title={Learning discrete world models for heuristic search},
  author={Agostinelli, Forest and Soltani, Misagh},
  jnournal={Reinforcement Learning Conference},
  year={2024}
}

@article{ha2018world,
  title={World models},
  author={Ha, David and Schmidhuber, J{\"u}rgen},
  journal={arXiv preprint arXiv:1803.10122},
  year={2018}
}

@article{oord2018representation,
  title={Representation learning with contrastive predictive coding},
  author={Oord, Aaron van den and Li, Yazhe and Vinyals, Oriol},
  journal={arXiv preprint arXiv:1807.03748},
  year={2018}
}

@article{grill2020bootstrap,
  title={Bootstrap your own latent-a new approach to self-supervised learning},
  author={Grill, Jean-Bastien and Strub, Florian and Altch{\'e}, Florent and Tallec, Corentin and Richemond, Pierre and Buchatskaya, Elena and Doersch, Carl and Avila Pires, Bernardo and Guo, Zhaohan and Gheshlaghi Azar, Mohammad and others},
  journal={Advances in neural information processing systems},
  volume={33},
  pages={21271--21284},
  year={2020}
}

@inproceedings{zbontar2021barlow,
  title={Barlow twins: Self-supervised learning via redundancy reduction},
  author={Zbontar, Jure and Jing, Li and Misra, Ishan and LeCun, Yann and Deny, St{\'e}phane},
  booktitle={International conference on machine learning},
  year={2021},
}

@article{meyer2023harnessing,
  title={Harnessing discrete representations for continual reinforcement learning},
  author={Meyer, Edan and White, Adam and Machado, Marlos C},
  journal={arXiv preprint arXiv:2312.01203},
  year={2023}
}

@article{lesort2018state,
  title={State representation learning for control: An overview},
  author={Lesort, Timoth{\'e}e and D{\'\i}az-Rodr{\'\i}guez, Natalia and Goudou, Jean-Franois and Filliat, David},
  journal={Neural Networks},
  volume={108},
  pages={379--392},
  year={2018},
  publisher={Elsevier}
}

@article{bagatella2021planning,
  title={Planning from pixels in environments with combinatorially hard search spaces},
  author={Bagatella, Marco and Ol{\v{s}}{\'a}k, Miroslav and Rol{\'\i}nek, Michal and Martius, Georg},
  journal={Advances in Neural Information Processing Systems},
  volume={34},
  pages={24707--24718},
  year={2021}
}

\end{document}